# Building-Road Collaborative Extraction from Remote Sensing Images via Cross-Task and Cross-Scale Interaction

Haonan Guo, *Student Member, IEEE*, Xin Su , *Member, IEEE*, Chen Wu, *Member, IEEE*, Bo Du, *Senior Member, IEEE*, and Liangpei Zhang, *Fellow, IEEE*.

*Abstract*—Buildings and roads are the two most basic man-made environments that carry and interconnect human society. Building and road information has important application value in the frontier fields of regional coordinated development, disaster prevention, auto-driving, etc. Mapping buildings and roads from very high-resolution (VHR) remote sensing images has become a hot research topic. However, the existing methods often extract buildings and roads with separate models, ignoring their strong spatial correlation. To fully utilize their complementary relation, we propose a method that simultaneously extracts buildings and roads from remote sensing images. The accuracy of both tasks can be improved using our proposed multi-task feature interaction and cross-scale feature interaction modules. To be specific, a multi-task interaction module is proposed to interact information across building extraction and road extraction tasks while preserving the unique information of each task. Furthermore, a cross-scale interaction module is designed to automatically learn the optimal reception field for buildings and roads under varied appearances and structures. Compared with existing methods that train individual models for each task separately, the proposed collaborative extraction method can utilize the complementary advantages between buildings and roads and reduce the inference time by half using a single model. Experiments on a wide range of urban and rural scenarios show that the proposed algorithm can achieve building-road extraction with outstanding performance and efficiency.

*Index Terms*—Building extraction, road segmentation, multitask learning, deep learning.

## I. Introduction

Buildings and roads are the two most basic man-made elements of our society. To be specific, buildings are basic carriers for social production and human life, while roads interconnect social network that enables goods and information transportation[1], [2]. The geodata of buildings and roads are valuable for understanding human activities and thus boost the sustainable development goals(SGD) including coordinated regional development, sustainable urban planning, public health, disaster risk reduction, etc.[3], [4]. Furthermore, the geoinformation of buildings and roads is indispensable for frontier applications such as smart cities and public health estimation[4], [5].

The very high-resolution (VHR) remote sensing imaging technique provides abundant image data sources for mapping buildings and roads in large areas[6]. Traditionally, these building and road maps are produced by visual interpreting, manual labeling, and surveying[7]. Despite its high accuracy, manually producing such geodata is time-consuming, especially for citywide or nationwide mapping[8]. Moreover, it cannot meet the requirements of up-to-date databases from

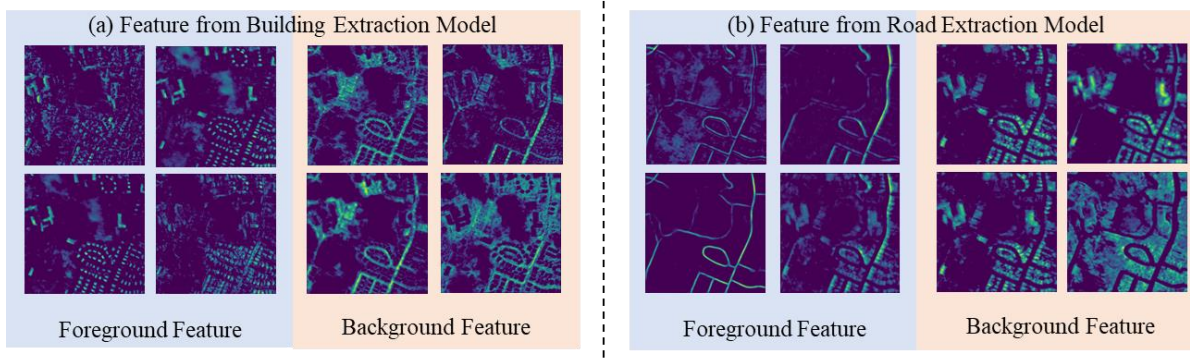

**Fig1.** Illustration of features from U-Net models that are trained individually on building extraction and road extraction tasks. Buildings and roads are complementary in the feature space.

This research was funded by the National Natural Science Foundation of China under Grant 42230108, 62371348 and 61801332. *Corresponding author: Xin Su* (E-mail: xinsu.rs@whu.edu.cn).

Haonan Guo, Chen Wu and Liangpei Zhang are with the State Key Laboratory of Information Engineering in Surveying, Mapping and Remote Sensing, Wuhan University, Wuhan, China (e-mail: haonan.guo@whu.edu.cn; chen.wu@whu.edu.cn; zlp62@whu.edu.cn).

Xin Su is with the School of Remote Sensing and Information Engineering, Wuhan University, Wuhan, China (e-mail: xinsu.rs@whu.edu.cn).
Bo Du is with the National Engineering Research Center for Multimedia Software, Institute of Artificial Intelligence, School of Computer Science and Hubei Key Laboratory of Multimedia and Network Communication Engineering, Wuhan University, Wuhan, China (e-mail: gunspace@163.com).

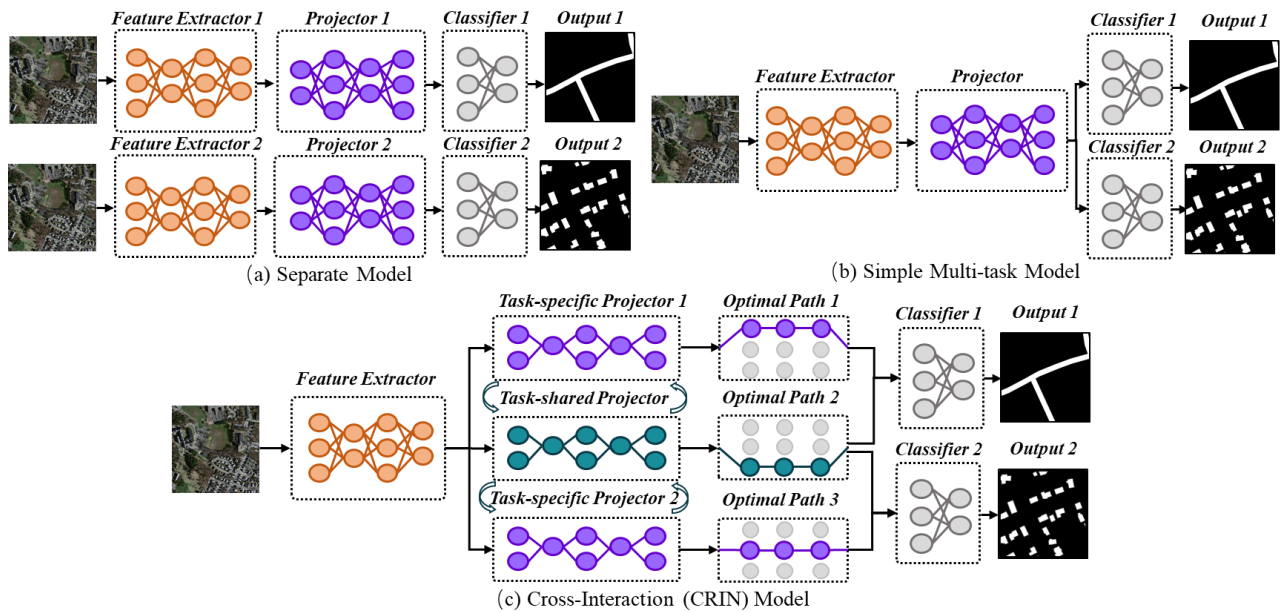

**Fig 2.** The comparison among different architectures for building-road collaborative extraction, including (a) separate models trained individually for different tasks (b) multitask model that predicts building and road using multiple classifiers, and (c) our proposed Cross-Interaction (CRIN) architecture that embeds task-shared and task-specific feature spaces and selective paths.

many geodata-oriented applications such as autonomous driving, since regular manual updating is laborious in the long run[9], [10], [11]. To this end, many studies have attempted to design automatic methods for mapping buildings and roads from VHR remote sensing images. Over the past decades, mainstream building and road extraction algorithms have developed from handcrafted feature-based methods to deep learning-based methods[12]. The traditional handcrafted-based methods seek to select spatial-spectral features that can well distinguish objects of interest from the background[13], [14], [15], [16]. However, limited by expert knowledge and feature generalization, it is time-consuming to manually design optimal feature descriptors or classifiers that can be generalized to large areas.

The convolution neural networks (CNN) of deep learning can automatically learn image descriptors and extract optimal features through backpropagation[17] and thus have been proven more effective than the handcrafted feature-based methods[18], [19]. Many studies have designed new CNN structures to improve the models' performance on buildings or roads[20], [21], [22]. However, despite the pivotal role that buildings and roads have in human society, many prevailing algorithms extract buildings or roads in isolation[23], [24], and ignore the strong spatial correlation relation between them. In Fig.1, we empirically find the strong correlation between building and road exists in the feature space when visualizing the features extracted by the building extraction model (Fig.1 a) and road extraction models (Fig.1 b) using the U-Net architecture[25]. An interesting finding is that some features extracted by a model trained only with building labels has learned to localize road regions. Similarly, the road segmentation model has automatically learned building information without supervision from the building ground truths. This finding implies that the building extraction and road extraction tasks can improve the accuracy of both tasks in a collaborative way since high-quality building features will boost the performance of the road extraction task and vice versa.

Although some existing methods have attempted to design unified network structures for building and road extraction, these models, as depicted in Fig. 2 (a), are trained on each task individually and cannot utilize the complementary relation between them[23], [24]. A question arises: can we design a multitask model for extracting buildings and roads simultaneously? An intuitive solution is to apply two classifiers for the two tasks and convert the model into a multi-task manner(Fig. 2 b) [26], [27], [28]. However, these simple multitask designs lead to the learning issue of the "seesaw phenomenon" that the accuracy of one task is often improved by hurting the performance of another task[29], which is also demonstrated in our experiment. Moreover, these methods do not consider the different requirements of models' reception fields between different tasks, which may also deteriorate multitask feature representation.

To address these problems, we propose a CRoss-INteraction (CRIN) model with a simple and effective task-shared and task-specific feature space design. As shown in Fig. 2(c), CRIN separates the common feature space into task-specific and task-shared feature spaces to realize inner-task and cross-task feature interaction, improving task-specific and task-shared feature representations. Moreover, as the optimal reception field of the different tasks varies, a cross-scale interaction module is designed to automatically learn and select the optimal reception field for each feature space, and thus exploit optimal feature representations for each specific task. The proposed CRIN model can boost the performance of both building extraction and road extraction tasks in a complementary way. Our main contributions are summarized as follows:

1. We propose a building-road extraction network that improves the accuracy of both tasks in a single model. A multitask interaction module is proposed to take advantage of the complement of spatial correlation relationship between buildings and roads. The proposed model can achieve cross-task interactions within the shared feature space while preserving the unique feature of each task under the inner-task interaction design. It alleviates the problem of isolated or unbalanced learning as proposed by the existing methods.

2. Since task-specific building and road features may depend on the model's reception field differently, a cross-scale interactive perception module is designed. The module adaptively determines the optimal receptive field for task-specific and task-shared feature spaces through back-propagation and thus improves the representation of task-specific and task-shared features.

3. Extensive experiments on a wide range of urban and rural scenes demonstrate the effectiveness of the method, which achieves the highest accuracy in building and road extraction tasks. In comparison to the existing structures that are trained separately or multitask structures that cause the 'seesaw phenomenon', the proposed CRIN model takes advantage of the complementary advantages between buildings and roads and achieves the best performance and efficiency.

## II. RELATED WORKS

### A. Building Extraction

Extracting buildings from VHR remote sensing images has long been a challenging task due to the high inner-class variation and small inter-class variation of VHR remote sensing images[30]. For example, non-building features with similar colors, shadows from nearby objects, and heterogeneity of roofs are some major problems that hinder the process of automation[14]. To overcome the confusion from non-building features, early methods seek to select features that can well represent buildings, and design multiple criteria to discriminate buildings from complex backgrounds[15], [31], [32]. As these criteria were designed manually, they usually show limited generalization ability when applied to large-scale applications[33].

In recent years, some studies introduced data-driven deep-learning techniques to the building extraction field. The advantage of deep learning is that it can automatically learn feature representation from the input image and make classifications in an end-to-end manner[17], [34]. Many works have been proposed to improve the fully convolutional networks(FCN) for building extraction. The main direction of improvements lies in building boundary refinement[35], [36], [37], multiscale feature fusion[38], [39], [40], and dilated convolution[30], [41]. There are also some studies that focused on applying multi-sensor images(e.g. LiDAR) [42], [43] or multi-source data(e.g. height information)[44], [45] for improved building extraction accuracy. Inspired by the outstanding performance of the Transformer mechanism[46], some recent studies have proven its efficacy in the building extraction field[47], [48]. Another research hotspot is the generalization of building vector maps. Unlike the semantic segmentation methods that generate pixel-level building prediction, building vectorization methods directly generate building polygons from images using post-processing[49], vertex prediction[50], or graph neural networks[51] methods. However, few studies have explored the performance gain of the model from learning objects with strong spatial correlation with buildings, such as roads.

### B. Road Extraction

Extracting roads from remote sensing images is also challenging due to diverse road materials and complex backgrounds[52]. Since roads are stripe-shaped objects with multiple intersections, spatial features such as geometric[53], topological[54], and texture[55] features are widely used in handcrafted feature-based methods. Similar to building extraction, the traditional handcrafted feature-based methods show limited performance when applied to large areas. The data-driven learning-based methods have shown better performance and generalization than traditional methods. Due to the distinctive shape characteristic of roads, traditional convolutional layers of small kernel size cannot well capture the shape information, and a mainstream research focus is to enlarge the models' reception field to capture road information[52]. For example, D-LinkNet introduces multiple branches of dilated convolutional layers to capture features at different scales[56]. As roads vary in length and materials, they rely on different levels of model reception fields. To further improve the model's reception field, transformer mechanisms were introduced to capture long-range dependency in images[57], and thus significantly improve the accuracy of road extraction[58], [59]. Based on the pixel-based road extraction results, some studies have attempted to convert the pixel-based road maps into polylines by designing vectorization approaches[60], [61]. To further improve road extraction accuracy, some studies also combined multi-source data such as vehicle trajectories[62] and SAR data[63]. However, it is still challenging to extract complete road map from images with complex background information.

### C. Building-Road Extraction

The existing building-road extraction methods can be categorized into separate methods and multitask methods. The separate methods aim to design a unified network architecture by considering the characteristics of buildings and roads. The network, however, is trained on building extraction and road extraction tasks separately. For example, Zhang and Wang proposed a unified network structure[24] with a large model reception field for building and road extraction. Ayala et al. introduced Res-U-Net to extract buildings and roads using crowdsource data as the training set[23]. Saito et al. proposed unified U-Net architectures embedded with recurrent neural networks and dense connections for building and road extraction[64]. These unified structures, however, do not consider the spatial correlation relation between buildings and roads. Moreover, these models are trained separately on different tasks, and thus double the computational burden since the image should pass through the model twice to generate building and road predictions. To this end, some studies designed multitask networks to predict the building and road in a single network. For example, some studies transmit the binary classification method into the multiclass classification method and classify the image patches into buildings, roads, and backgrounds [27], [64], [65]. These patch-based methods do not utilize the global context information and are prone to cause large building misclassification and road disconnection due to insufficient spatial semantics[64]. Some further studies introduce FCN to extract buildings and roads in a single model. Ding et al. proposed a non-local feature search network [28] embedded with a self-attention module to strengthen feature representation. Ayala et al. [66] designed multitask strategies to train each category against the background before and fuse the predictions using post-processing.

There are two major issues that hinder the performance of the building-road extraction, including 1) many existing multitask learning schemes simply attach several classifiers to generate multiple predictions, which may lead to the 'seesaw phenomenon' that features of one task become dominant and sacrifice the performance of the other task, and 2) The model reception fields for building and road extraction tasks are unified, ignoring the different requirements of model reception fields for different tasks. In this paper, we overcome these shortcomings by separating the common feature space into task-specific features and task-shared future spaces with inner-task and cross-task feature interaction. A cross-scale interaction module then is designed to automatically learn and select the optimal reception field for both task-shared and task-specific feature spaces.

## III. METHODOLOGY

### A. Overview

To take advantage of the correlation relationship between buildings and roads, we propose a CRoss-INteraction (CRIN) network to realize the simultaneous extraction of buildings and roads in a single network. As shown in Fig. 3, CRIN takes the VHR remote sensing images as the input and outputs the prediction of building and road maps simultaneously to improve learning efficiency through information sharing between tasks. Following the encoder-decoder design, the input images are first fed into the encoder network for hierarchical feature extraction. The extracted features are then copied and transmitted to the decoder network to refine feature resolution and generate building and road extraction. In each decoder stage, a multitask feature interaction (MFI) module is designed to address the 'seesaw phenomenon' that deteriorates multitask learning, which includes the inner-task fusion stage and the cross-task interaction stage. MFI performs inner-task feature fusion to integrate useful information coming from the encoder network; a cross-task interaction scheme then separates the feature space into task-specific and task-shared feature spaces for better feature interaction. In the task-specific feature space, the features are expected to learn representations that are optimal for the specific task, while the features of both tasks interact in the shared feature space. As a result, MFI is able to preserve the features of each specific task and take advantage of the complementary between building features and road features. Furthermore, a cross-scale interaction (CSI) module is integrated after MFI to select the optimal feature scale for each task, since different objects of interest rely on the reception field differently. Finally, the model is optimized with the building and road ground truths to handle the practical challenges of joint learning.

### B. Multi-task interaction module

Motivated by the strong spatial correlation relationship between buildings and roads, we expect to train a model that utilizes the complementary advantages between buildings and roads and outperforms the single-task model that is trained separately. We propose a multi-task interaction (MTI) module to fully utilize the complementary advantage between buildings and roads. As shown in Fig. 4, the proposed MFI tasks the output features from the last decoder stage and shallow features transmitted from the encoder network as the input. Deep features are rich in contextual semantics, but their edge details are insufficient due to low spatial resolution. On the contrary, features extracted by shallow encoder convolutional layers contain rich edge details of the whole image other than the objects of interest. In conventional decoder design, the common is to concatenate the multiple features and fuse them using a single convolutional feature. Another convolutional layer then refines the features and outputs the feature to the next decoder stage. However, in the multitask learning scheme, directly concatenating and convoluting multiple features into a unified representation leads to the 'seesaw phenomenon' in which the features of one task become dominant and sacrifice the performance of the other task[29].

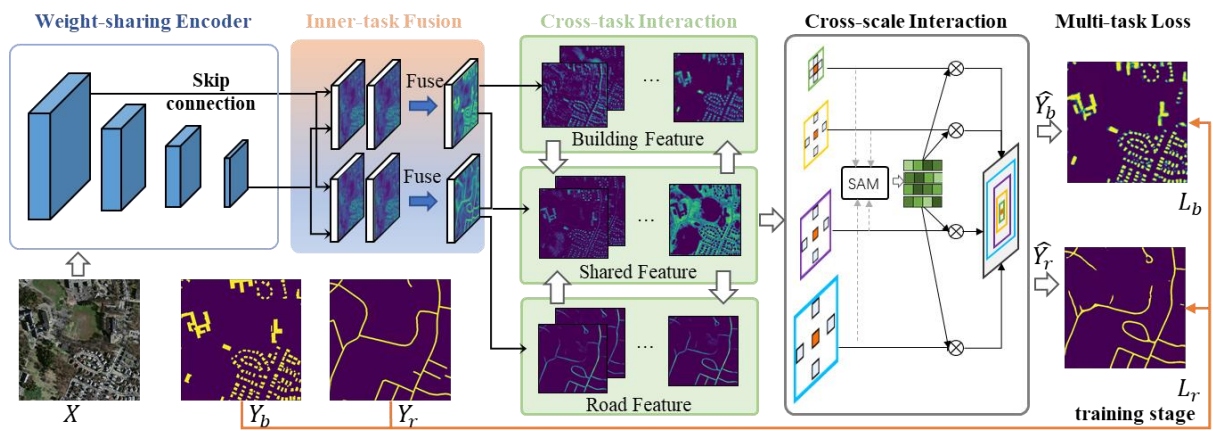

**Fig 3.** The flowchart of the proposed Cross-Interaction (CRIN) network, which includes a weight-sharing encoder network, a multitask interaction module (MFI) that enables inner-task and cross-task feature interaction, and a cross-scale interaction module (CSI) to select and integrate optimal feature scales.

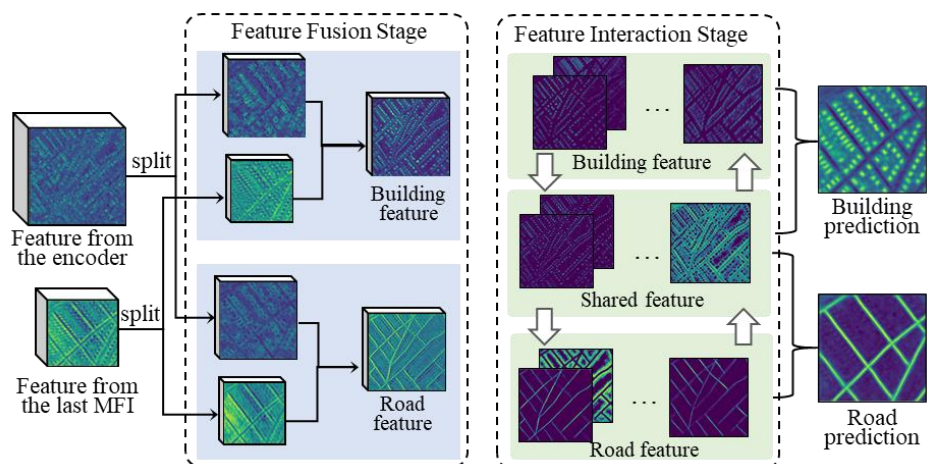

**Fig 4.** Architecture of the multi-task interaction (MTI) module.

To address this problem, we aim to separate the unified feature space to ensure a proper learning space for each task, which also allows information sharing. The proposed MTI module includes the feature fusion stage and the feature interaction stage, as illustrated in Fig. 4. In the fusion stage, we guide the model fuse building features and road features separately since different tasks focus on different details contained in the input feature. We split the input features into two groups and concatenate features from each group in alternation. Convolution with a group size of two is performed to the concatenated features to generate building features and road features that contain rich semantic and spatial details. In this way, features from each feature space can focus on refining the spatial details of the specific task. The real-world example in Fig. 4 shows that the fusion result of each task can better locate building and road regions compared with the input feature. The features are then projected for cross-task interaction in the interaction stage.

In the interaction stage, we further project the features into the task-specific feature spaces, including the building feature space and the road feature space, and a task-shared feature space that enables cross-task feature interaction. In this way, features of both tasks can interact in the shared feature space, and task-specific features can be preserved in the task-specific feature space.

We then explain why MFI is a lightweight and effective decoder design for building-road collaborative extraction. If we directly adopt the conventional U-Net decoder for collaborative extraction, one task may become dominant in the training process, and the features cannot be well balanced between different tasks[29]. However, in MFI, we separate the decoder into the feature fusion stage and the feature interaction stage, where the former is responsible for fusing the features for each task and the latter is capable of information interaction across the different tasks. Moreover, MFI can be implemented by successive feature splitting and group convolutions, which saves one-third of the computational cost as compared with the conventional decoders in theory. It demonstrates that MFI is an effective and lightweight decoder module for building-road collaborative segmentation.

### C. Cross-scale interaction module

MFI separates task-specific and task-shared features explicitly, and thus preserves the feature space of each task and allows cross-task feature interaction. As depicted in Fig. 5, CSI contains multiple branches of different large convolutional kernels, which refer to different scales of the reception field. Large kernel sizes with depthwise convolutional layers have been proven effective in improving the model's reception field. However, since the optimal reception field for different tasks varied, manually selecting a scale of large kernel size may not be helpful in improving multitask accuracy. To address this problem, the input features are fed into multiple branches of different large convolutional kernels and introduce a selective kernel mechanism to determine the contribution of each scale. Moreover, to alleviate the heavy computational burden of multiple large-kernel branches, we introduce alternative row-wise convolution and column-wise convolution, which effectively reduces the computational cost and improves the receptive field of the model.

Specifically, CSI takes the output feature f of MTI as the input. f is firstly fed into a depthwise convolutional layer of kernel size of 5x5 to extract the initial feature $f_{init}$:

$$f_{init} = DConv_{5\times5}(f) \quad (1)$$

in which $DConv_{i\times j}(\cdot)$ represents a depthwise convolutional layer of kernel size ixj. Then $f_{init}$ is passed through 4 branches including the reception field of 7x7, 11x11, 21x21, and the residual branch that maintains the original reception field. Moreover, if we directly feed the features into convolution layers of kernel size kxk, the model's parameters increase exponentially. For example, the parameters of a convolution layer of 21x21 kernel size are 49 times that of the 3x3 kernels. Inspired by the inception module, we apply column-wise and row-wise convolution in success to increase the model reception field with less computational cost. The computation process of these branches is as follows.

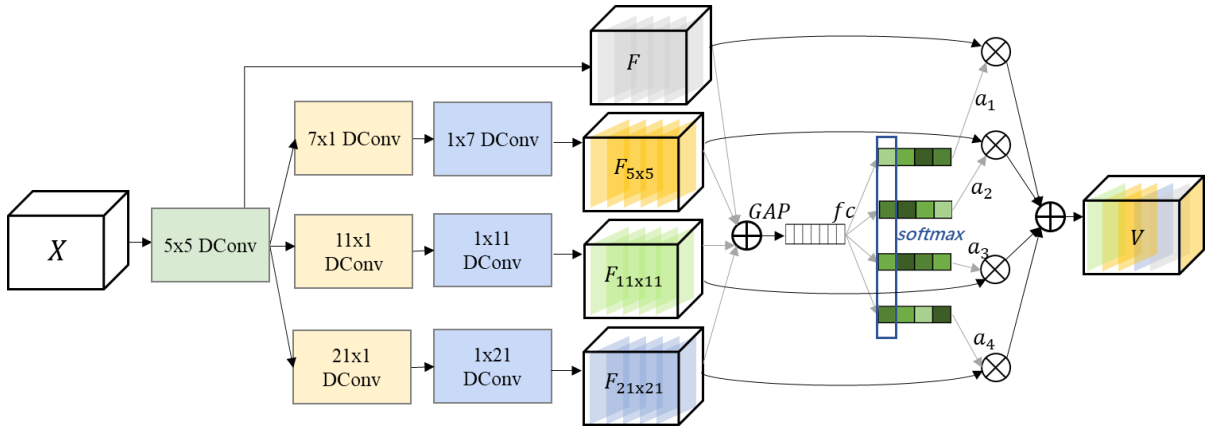

**Fig 5.** The architecture of the cross-scale interaction (CSI) module.

$$f_{7x7} = \text{Conv}_{1x7}(\text{Conv}_{7x1}(f_{init})) \quad (2)$$

$$f_{11x11} = \text{Conv}_{1x11}(\text{Conv}_{11x1}(f_{init})) \quad (3)$$

$$f_{21x21} = \text{Conv}_{1x21}(\text{Conv}_{21x1}(f_{init})) \quad (4)$$

$$f_{skip} = f_{init} \quad (5)$$

Given the features of different reception fields, the next step is to decide the contribution of each branch, which is the key step in determining the optimal scale for each task. We introduce a scale attention module to predict the contribution of each scale. Features of all scales are summed pointwise, followed by a global average pooling module that aggregates global context information of all scales. A multilayer perceptual layer is connected to generate a vector of shape 4xC, followed by a softmax operation that converts the output into scale attention values, where 4 represents the attention value of each scale. This contribution value is then multiplied by the features of each scale along the channel axis. The final output of CSI is the summation of all the attention-activated features:

$$f_{fuse} = GAP(f_{7x7}) + GAP(f_{11x11}) + GAP(f_{21x21}) + GAP(f_{init}) \quad (6)$$

$$Attn = softmax(MLP(f_{fuse})) \quad (7)$$

$$f_{out} = Attn_1 \cdot f_{7x7} + Attn_2 \cdot f_{11x11} + Attn_3 \cdot f_{21x21} + Attn_4 \cdot f_{init}, where \sum_{i}^{i=4} Attn_i = 1 \quad (8)$$

We then explain why CSI is capable of learning the optimal scale for different tasks. Since we adopted depthwise convolution in the multi-branch stages and the scale attention stage, the optimal scale for different feature layers is learned independently. It means that features from different tasks can automatically learn the contribution of each scale without inference from the other task. Compared with many existing works that design a unified framework for buildings and roads in ignorance of different requirements of the reception field, CSI can automatically determine the optimal scale of each task and extract features with the optimal reception field. Another advantage of CSI is that it is a super lightweight module that can effectively improve the reception field for different tasks. By introducing depthwise 1xn and nx1 convolutions in succession, the computation burden is reduced to 1/600 compared with CSI without these improvements.

### D. Model Details

The CRIN model is implemented under the deep learning framework of Pytorch and is optimized using the AdamW optimizer[67]. The learning rate is set to 0.001 initially and is adjusted under the 'poly' scheduler with power 0.9. The model is trained for 40k iterations with a batch size of 16. To avoid overfitting, we adopted a data augmentation strategy that randomly rotates, flips, and scales the input image and label patches. The loss function of CRIN includes three parts, the building segmentation loss, the road segmentation loss, and the deep supervision loss. The building segmentation loss and the road segmentation loss are the summations of cross entropy loss and dice loss between the building and road prediction $\widehat{y_b}, \widehat{y_r}$ and their corresponding ground truths $y_b$ and $y_r$:

$$L_{building} = (\text{Dice}(\widehat{y_b}, y_b) + \text{CE}(\widehat{y_b}, y_b))/2 \quad (9)$$

$$L_{road} = (\text{Dice}(\widehat{y_r}, y_r) + \text{CE}(\widehat{y_r}, y_r))/2 \quad (10)$$

We also introduce the deep supervision loss to guide the model to learn task-specific and task-shared features. Given the task-specific building feature $f_b$, road feature $f_r$, and the task-shared feature $f_s$, we apply deep supervision to generate deep building and road predictions from these features. Specifically, the prediction of each task is generated by the task-specific feature and task-shared feature:

$$L_{aux} = \sum_{i=1}^{n} \text{CE}(Conv_{1x1}([f_b^i, f_s^i]), y_b) + \text{CE}(Conv_{1x1}([f_r^i, f_s^i]), y_r) \quad (11)$$

In this way, features from task-specific feature space are able to learn information that is helpful to the specific task, and the task-shared features are able to learn common features that are beneficial to both tasks. The overall loss is the weighted summation of the building segmentation loss, the road segmentation loss, and the deep supervision loss. Since the deep supervision loss is the auxiliary loss that guides model learning, we adopted a weight of 0.1 to $L_{aux}$, and the final loss function is calculated as follows:

$$L = L_{building} + L_{road} + 0.1 * L_{aux} \quad (12)$$

After the model converges, the model's performance is evaluated on the test dataset.

## IV. EXPERIMENTS

### A. Datasets

We select two large-scale building and road datasets to evaluate the performance of the proposed method.

The Massachusetts dataset[65]: The Massachusetts building and road dataset is a large-scale dataset that covers a wide variety of urban and rural regions in Massachusetts. The building and road labels are generated by the OpenStreetMap (OSM) project. Each image has an image size of 1500x1500 with a spatial resolution of 1m. Following the data partition method in [64], we collect the patches that contain both building and road labels and clip them into 512x512 patches with a stride ratio of 0.5. As a result, the dataset includes 3077, 200, and 250 patches for training, validation, and testing, respectively.

The aerial imagery object identification dataset[68]: The aerial imagery object identification(AIOI) dataset is a multi-city dataset, which contains images and labels from 9 different cities including urban and suburban areas. This dataset is suitable for evaluating the model's performance in large-scale and cross-city mapping. The image resolutions range from 0.15 to 0.3 meters; the ground truth labels are produced using OSM. The dataset includes images from suburban areas of 6 cities and urban areas from 3 cities. We randomly selected New York, Atlanta, and New Heaven as the test dataset and the rest as the training dataset. As a result, the dataset includes 9806 patches for training and 2840 patches for testing. The cross-city mapping of the AIOI dataset poses a great challenge to the generalization and performance of the models.

### B. Evaluation Metrics

The models' performance is evaluated by calculating the precision, recall, F1-score, and IoU between the model predictions and the ground truths. Precision represents the ratio of predicted foreground pixels that are correctly predicted; recall indicates the proportion of the ground truth buildings or roads that are correctly predicted; F1-score is the harmonic average between precision and recall. IoU denotes the intersection between prediction and ground truth over their union. These evaluation metrics can be calculated as follows.

$$\text{Precision} = \frac{TP}{TP+FP} \tag{13}$$

$$\text{Recall} = \frac{TP}{TP+FN} \tag{14}$$

$$\text{F1 Score} = 2 * \frac{Precision*Recall}{Precision+Recall} \tag{15}$$

$$\text{IoU} = \frac{TP}{TP+FN+FP} \tag{16}$$

where TP, FP, and FN represent the number of true positive, false positive, and false negative pixels in the test dataset, respectively.

Furthermore, the implementation speed and the computational cost of the model are important in large-scale mapping scenarios, thus we evaluated the model complexity using three metrics including the amount of network parameters, floating point operations (FLOPs), and frame per second (FPS). The parameter number reflects the model size; FLOPs denote the computational cost of the model; FPS corresponds to the inference speed of the model.

### C. Comparison Methods

We compare our proposed method with seven building-road collaborative extraction methods. These comparison methods can be categorized into separate methods[23], [24], [26], [65] and multitask methods[27], [28], [64]. The separate methods are unified architectures that are trained on building and road extraction tasks individually; the multitask methods introduce multiple classifiers to generate building and road predictions in a single model. We give a brief introduction to these methods as follows.

➢ Mnih and Volodymyr[65] proposed a patch-based classification network to extract buildings and roads in separation. The patches of 64x64 pixels are cropped from the image and predict the category of the central patches of 16x16 pixels. The models are trained and inferred individually with building and road extraction tasks.

TABLE I QUANTITATIVE COMPARISON WITH SOTA METHODS ON THE MASSACHUSETTS DATASET

| Method | Type | Task | IoU | Precision | Recall | F1-Score |
|---|---|---|---|---|---|---|
| **CRIN(ours)** | | Building | **71.15** | 83.79 | **82.5** | **83.14** |
| | | Road | **59.09** | 75.49 | 73.12 | **74.29** |
| Saito et al. | Multitask | Building | 64.17 | 80.64 | 75.86 | 78.17 |
| | | Road | 53.47 | 73.26 | 66.43 | 69.68 |
| Alshehhi et al. | | Building | 65.36 | **85.8** | 73.29 | 79.05 |
| | | Road | 56.27 | **82.98** | 63.61 | 72.02 |
| NFSNet | | Building | 59.11 | 76.14 | 72.56 | 74.3 |
| | | Road | 51.88 | 70.19 | 66.55 | 68.32 |
| JointNet | Separate | Building | 69.38 | 85.52 | 78.61 | 81.92 |
| | | Road | 56.89 | 79.11 | 66.95 | 72.52 |
| MCG-UNet | | Building | 66.59 | 84.45 | 75.9 | 79.94 |
| | | Road | 57.79 | 76.4 | 70.35 | 73.25 |
| Res-U-Net | | Building | 67.44 | 78.88 | 82.31 | 80.56 |
| | | Road | 56.98 | 76.62 | 68.96 | 72.59 |
| Mnih et al. | | Building | 62.64 | 81.9 | 72.7 | 77.03 |
| | | Road | 53.22 | 75.95 | 64 | 69.47 |

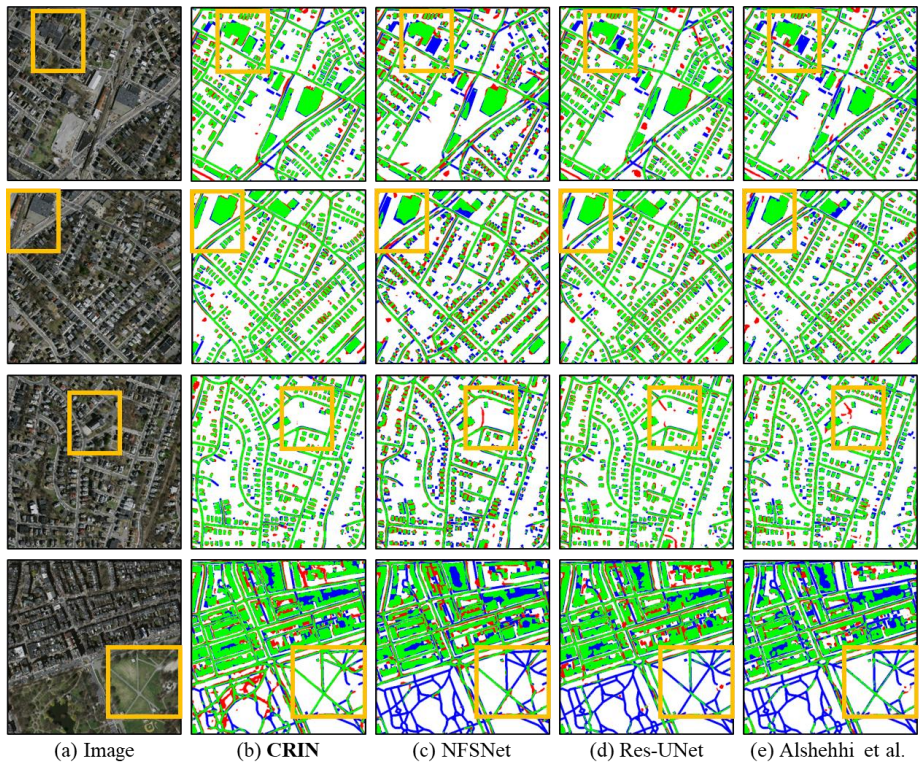

**Fig 6.** Visualization of the building-road extraction results of the comparison methods in the Massachusetts dataset. Pixels in green, blue, and red are correctly predicted, omitted, and misclassified, respectively.

- Ayala et al.[23] proposed a modified version of U-Net using ResNet as the encoder network. This model is utilized to extract buildings and roads in separation. The predictions are merged after inferring the building and road predictions separately.

- Abdollahi et al.[26] introduce novel mechanisms, including bi-directional LSTM, dense connection, and attention modules to improve the model performance on building and road extraction. The model is trained on building and road extraction tasks in separation. We select the reported best MCG-Net for comparison.

- JointNet[24] is a unified and effective architecture for building and road extraction. The network combines atrous convolution with densely connected convolutions to enlarge the model reception field. The model is trained separately on building and road extraction tasks.

- The non-local feature search network(NFSNet) can extract buildings and roads in a single model that generates multiple channels in the output prediction[28]. Self-attention modules and feature refinement modules are designed to alleviate building misjudgment and road disconnection problems.

- The method as proposed by Alshehhi et al. is a patch-based method for the simultaneous extraction of buildings and roads[27]. The category of the central 16x16 pixels is predicted from the surrounding 64x64 pixels. A post-processing method is introduced to alleviate the misclassification problem using superpixels.

- Saito et al. proposed a multiple-object extraction method for building and road extraction[64]. It is a patch-based method that introduced channel-wise inhibited softmax. The final prediction is generated patch-by-patch under model averaging.

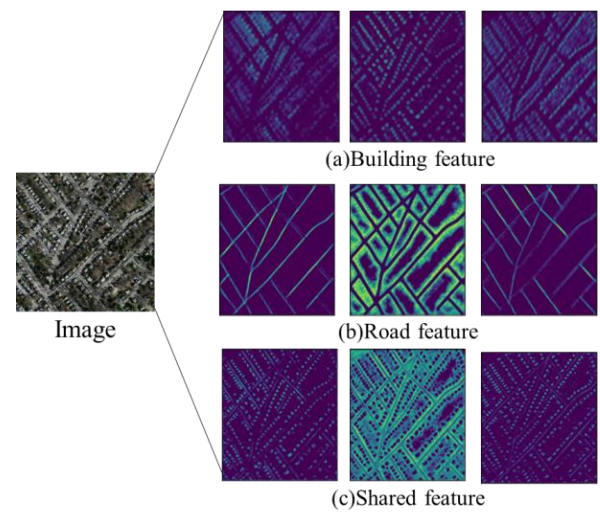

**Fig 7.** Visualization of the features in the task-specific and task-shared feature spaces.

TABLE II ABLATION EXPERIMENTAL RESULTS ON THE MASSACHUSETTS DATASET

| Method | Multitask | MTI | CSI | Task | IoU | Precision | Recall | F1-Score | GFlops |
|---|---|---|---|---|---|---|---|---|---|
| **Baseline** | | | | Building | 69.28 | 81.73 | 81.98 | 81.85 | 1.57×2 |
| | | | | Road | 55.07 | 75.09 | 67.38 | 71.03 | |
| Ablation 1 | √ | | | Building | 69.16 | 85.19 | 78.62 | 81.77 | 1.62 |
| | | | | Road | 57.69 | 76.73 | 69.92 | 73.17 | |
| Ablation 2 | √ | √ | | Building | 70.14 | 85.56 | 79.56 | 82.45 | 0.91 |
| | | | | Road | 57.86 | 78.65 | 68.65 | 73.31 | |
| Ablation 3 | √ | √ | √ | Building | 71.15 | 83.79 | 82.5 | 83.14 | 0.96 |
| | | | | Road | 59.09 | 75.49 | 73.12 | 74.29 | |

Table III Quantitative Comparison with SOTA Methods on the AIOI dataset

| Method | Type | Task | IoU | Precision | Recall | F1-Score |
|---|---|---|---|---|---|---|
| **CRIN(ours)** | | Building | **66.25** | **79.99** | **79.41** | **79.7** |
| | | Road | **52.22** | **68.98** | **68.25** | **68.61** |
| Saito et al. | Multitask | Building | 38.02 | 60.24 | 50.75 | 55.09 |
| | | Road | 18.15 | 35.9 | 26.86 | 30.73 |
| Alshehhi et al. | | Building | 38.12 | 58.37 | 52.35 | 55.2 |
| | | Road | 17.93 | 31.96 | 28.99 | 30.4 |
| NFSNet | | Building | 60.4 | 77.82 | 72.96 | 75.31 |
| | | Road | 48.11 | 70.24 | 60.42 | 64.96 |
| JointNet | | Building | 47.5 | 71.74 | 58.43 | 64.41 |
| | | Road | 30.27 | 56.07 | 39.67 | 46.47 |
| MCG-UNet | Separate | Building | 51.27 | 76.12 | 61.1 | 67.79 |
| | | Road | 38.3 | 67.25 | 47.08 | 55.38 |
| Res-U-Net | | Building | 57.07 | 77.21 | 68.63 | 72.67 |
| | | Road | 47.2 | 71.06 | 58.43 | 64.13 |
| Mnih et al. | | Building | 37.48 | 67.88 | 45.55 | 54.52 |
| | | Road | 21.99 | 33.73 | 38.73 | 36.06 |

### D. Experiments on the Massachusetts dataset

The Massachusetts building and road dataset is a large-scale dataset that covers a wide variety of urban and rural regions in Massachusetts, US. The experimental results of CRIN and the comparison methods are presented in Table I, with the highest score marked in bold and the second highest score underlined. From Table I we can see that our proposed CRIN achieves the IoU and F1-score on both building and road extraction tasks. Although the method proposed by Alshehhi et al. performs well in the precision score, the recall score is rather low, indicating that many foreground pixels are omitted. CRIN outperforms the separate networks by utilizing the spatial correlation between buildings and roads, thus improving the accuracy of both tasks. Among the multitask methods, the methods as proposed by Saito et al. and Alshehhi et al. are the patch-based methods that cannot fully utilize the spatial details from VHR remote sensing images, since only the 64x64 patches are fed forward the model. The segmentation-based NFSNet, however, cannot perform well on the Massachusetts dataset since it generates predictions from features that obtain 1/16 of the image resolution, causing serious omission of tiny buildings and roads. On the contrary, our proposed CRIN model is a multitask model that obtains accurate building and road extraction from remote sensing images.

We visualize some examples of the building and road predictions in Fig. 6, where the green, blue, and red pixels represent the correctly predicted, the omitted, and the misclassified pixels. From Fig. 6 rows 1-2, we can see that NFS-Net and the method proposed by Alshehhi et al. omitted some large-scale buildings nearby roads. However, CRIN can preserve the integrity of the predicted buildings by considering the spatial correlation between buildings and roads. Moreover, as shown in Fig. 6 rows 3-4, some tiny roads are misclassified or omitted by all the comparison methods, but CRIN can correctly predict the tiny roads by adaptively selecting the optimal reception field for perceiving the roads. CRIN is an effective model for simultaneously extracting buildings and roads from remote sensing images.

Furthermore, we conducted ablation experiments on the Massachusetts dataset to validate the effectiveness of the proposed modules. We selected the U-Net architecture with the EfficientNet backbone as the baseline. This model is trained and inferred separately on the building and road extraction tasks. From Table 2, we can see that the baseline model achieved the F1-score of 81.85% and 71.03% on building extraction and road extraction tasks. If we simply add multiple classifiers to the baseline and convert it into a multitask model, the accuracy of the road extraction task is improved by 2.14% on the F1-score. However, the model's performance on building extraction is decreased. It demonstrates the 'seesaw phenomenon' of multitask learning where the road extraction task is improved with the building extraction task sacrificed. This is because the road extraction task becomes dominant in the training process and thus leads to unbalanced learning. If we improve multitask learning by integrating the MTI module, the accuracy of both tasks can be improved by fully utilizing the spatial correlation between buildings and roads. As a result, the F1-score of buildings and roads are improved by 0.6% and 2.28%, respectively. Furthermore, if we utilize the CSI module to learn the optimal reception field for different tasks, the accuracy of both tasks can be further improved.

We also present the model complexity in Table 2 using GFlops. The baseline model has the highest complexity because the model is inferred twice to generate the building and road predictions. The complexity of a simple multitask network is appropriate half of the baseline model. It should be noted that the model complexity is obviously decreased with MTI as compared with the multitask model, which is because MTI reduces the computation cost using successive group convolution. Furthermore, CSI slightly increases the computational cost and achieves the best performance on both tasks.

The intermediate features of the task-specific and task-shared feature space of the MTI module are presented in Fig.7. From Fig. 7 (a) and (b) we can see that the task-specific building and road features focus on localizing buildings and roads respectively. Meanwhile, features from the shared feature space focus on capturing the context information and co-location pattern between buildings and roads. The accuracy of buildings and roads can be improved by separating task-specific and task-shared features.

E. *Experiments on the AIOI dataset*

We conducted experiments on another large-scale AIOI dataset. The AIOI dataset includes images and labels from 9 cities that cover various urban and suburban scenes, which is suitable for evaluating the generalization of the proposed method. As the training dataset and test dataset are image patches from different cities, the models' performance of large-scale mapping in practical scenarios can be testified. The quantitative comparison as listed in Table III shows that the CRIN model outperformed the comparison models and achieved the highest accuracy in both buildings and roads. The patch-based methods as proposed by Saito et al. and Alshehhi et al. cannot perform well on large-scale mapping because these methods cannot fully utilize the spatial information of VHR remote sensing images. Meanwhile, although limited performance on the Massachusetts dataset, NFSNet achieves the second highest score on the AIOI dataset, which indicates the performance of NFSNet is influenced by the spatial resolution of images.

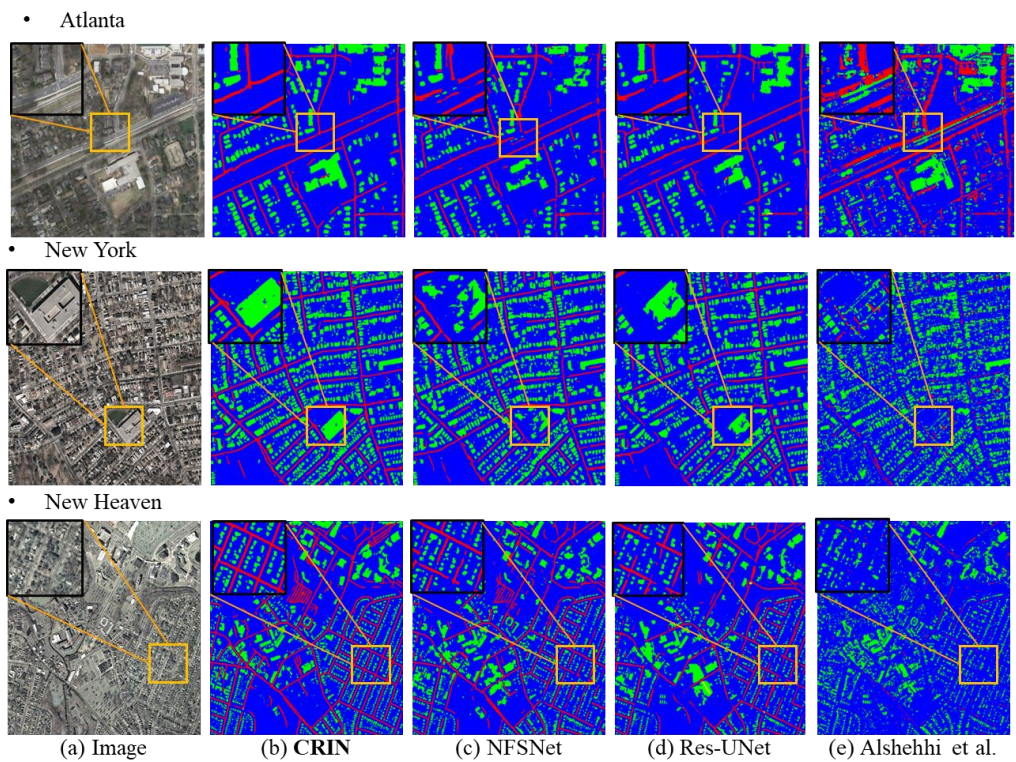

**Fig 8**. Visualization of the building and road extraction results on the AIOI dataset.

TABLE IV ABLATION EXPERIMENTAL RESULTS ON THE AIOI DATASET

| Method | Multitask | MTI | CSI | | IoU | Precision | Recall | F1-Score | GFlops |
|---|---|---|---|---|---|---|---|---|---|
| **Baseline** | | | | Building | 61.93 | 79.21 | 73.95 | 76.49 | 1.57×2 |
| | | | | Road | 48.4 | 66.37 | 64.13 | 65.23 | |
| Ablation 1 | √ | | | Building | 60.65 | 79.51 | 71.88 | 75.5 | 1.62 |
| | | | | Road | 48.8 | 67.05 | 64.2 | 65.59 | |
| Ablation 2 | √ | √ | | Building | 62.93 | 80.04 | 74.64 | 77.25 | 0.91 |
| | | | | Road | 49.52 | 65.99 | 66.48 | 66.24 | |
| Ablation 3 | √ | √ | √ | Building | 66.25 | 79.99 | 79.41 | 79.7 | 0.96 |
| | | | | Road | 52.22 | 68.98 | 68.25 | 68.61 | |

The ablation experimental results as shown in Table IV demonstrate the effectiveness of the proposed MTI and CSI modules. The 'seesaw phenomenon' also leads to an accuracy drop in the building extraction task of the multitask method. The proposed MTI enables balanced learning and thus improves the model accuracy of both tasks. The CSI module further improves the model performance on different tasks by learning the optimal reception field. As a result, our proposed CRIN model improves the F1-score by 3.21% and 3.38% on building and road extraction tasks respectively with less computational cost, demonstrating the model's effectiveness in large-scale and cross-city mapping.

We visualize the large-scale mapping results of different cities in Fig. 8, including Atlanta, New York, and New Heaven. From Fig. 8 we can see that CRIN obtains the best qualitative visualization results as compared to the other methods. The buildings and roads are seriously omitted by the patch-based method. Meanwhile, some buildings and roads as predicted by NFSNet and Res-UNet are incomplete and discontinuous. On the contrary, CRIN successfully depicts the buildings and roads by taking advantage of the complementary between buildings and roads. CRIN also outperforms the comparison methods by precisely depicting buildings and roads from suburban areas, as shown in Fig.8 row 3. The quantitative and qualitative experimental results demonstrate the generalization ability of the proposed CRIN method in various urban and rural areas.

## V. DISCUSSION

In this section, we will discuss and compare the complexity of models, visualize, and analyze whether different tasks rely on the same model reception field, and discuss the limitations of our proposed method.

TABLE V MODEL COMPLEXITY OF THE COMPARISON METHODS

| Method | Type | GFlops | Parameters | FPS |
|---|---|---|---|---|
| **CRIN** | | 0.96 | 4.38 | 38.51 |
| Saito et al. | Multitask | 144.27 | 2.42 | 1.02 |
| Alshehhi et al. | | 41.8 | 19.45 | 1.59 |
| NFSNet | | 9.82 | 11.91 | 63.3 |
| Joint-Net | Separate | 417.93 | 8.84 | 9.52 |
| MCG-UNet | | 1325 | 55.78 | 7.68 |
| Res-U-Net | | 83.74 | 50.2 | 28.85 |
| Mnih et al. | | 79.31 | 34.71 | 0.6 |

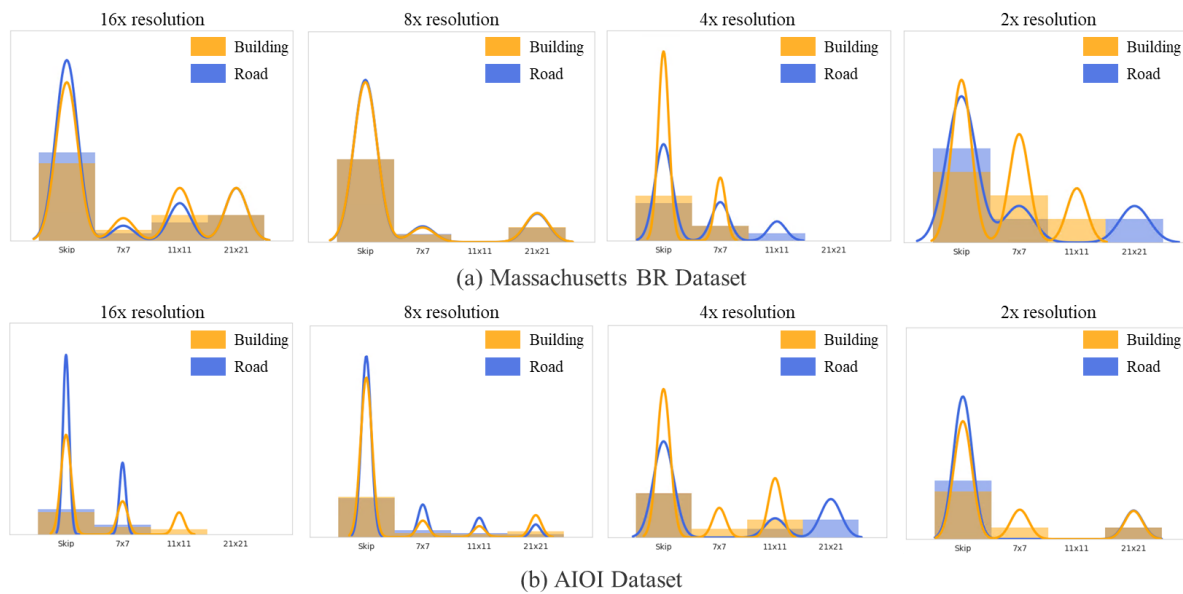

**Fig 9.** The contribution of different scales on different feature resolutions and datasets.

## A. Model Complexity

Apart from model accuracy and generalization ability, the complexity of models also matters in practical applications. In particular, the number of parameters and flops reflects the computational cost of the model in the inference stage; the frame per second (FPS) index indicates the time used for inference. Table V shows the number of parameters, flops, and FPS of each model testified on a single RTX 3090 GPU. In general, the separate models infer slower than the multitask model with higher Flops and lower FPS since these models are trained separately on roads and buildings, and need to be inferred twice to obtain buildings and roads prediction. As a result, the number of flops and parameters is doubled as compared to the single model. The Flops of Joint-Net and MCG-UNet are rather high since these methods maintain high feature resolution with multiple channels in the model, which leads to high computational costs. Among the multitask methods, the proposed CRIN obtains the lowest flops and the second highest FPS, which indicates the low computational cost and high inferring speed of our proposed methods. In comparison, the methods proposed by Saito et al. and Alshehhi et al. are much slower by approximately inferring a single 512x512 image per second. The reason is twofold. These patch-based methods receive 64x64 image patches as the input, and clipping these image patches with a stride of 16x16 leads to multiple inferring times as compared with the FCN-based methods. Moreover, to capture context information, these patch-based methods require input patches that are 4-time larger than the output prediction, resulting in redundant computation in the inference stage. In comparison, the FCN-based CRIN and NFS-Net are faster than not only the separate methods but also the patch-based methods. However, despite high FPS, NFSNet is prone to omit small buildings and slim roads since the predictions are generated from very low-resolution features (16 times lower than the input image). CRIN, on the contrary, can balance computation cost, inference speed, and accuracy.

## B. Scale contribution

In the proposed CSI module, multiple branches are designed to extract features by different reception fields, including skip connection, 7x7, 11x11, and 21x21; a scale attention module then determines the contribution of each scale and fuses the features based on the attention value. The CSI module can determine the optimal reception field for the task-shared and task-specific feature spaces by applying the argmax operation to each channel in the features. Moreover, the CSI module is integrated into different stages of the decoder module and can determine the optimal reception field at different feature resolutions. A question arises: do task-specific building and road features rely differently on the reception field and how?

In Fig. 9, we visualize the optimal receptive field of the task-specific feature spaces on different resolutions and datasets. From Fig. (a), we can see that the optimal reception field for buildings and roads is similar, with most features remaining the original reception filed by skip connection, and some features select large convolution kernels such as 11x11 and 21x21 as the optimal reception fields. It indicates that the high-level feature presentations for buildings and roads rely similarly on the model reception field. However, as the feature resolution increases, the optimal feature reception field for different tasks differs. More specifically, the building extraction task relies more on smaller reception fields such as skip connection, and 7X7; the road extraction task relies heavier on large reception fields such as 21x21 and 11x11 than the building extraction task. This is probably because roads are objects of linear shapes and depend on the consistency of shape and position. As a result, fusing shallow features from the encoder network plays an important role in recovering spatial information[24] and thus the optimal reception field of the road extraction task is larger to perceive and recover spatial information from high-resolution features. From Fig.9 (b) we can see that the optimal receptive field 21x21 for road extraction appears in the 4x resolution scale, which is 2 times lower than the Massachusetts dataset. Considering that the spatial resolution of the Massachusetts dataset is approximately 3 times lower than the AIOI dataset, we suppose that the optimal reception field of different tasks is related not only to the depth of the decoder network but also to the image spatial resolution. Designing (or searching) the optimal network architecture for different feature scales or resolutions is an interesting and open question to investigate in the future.

### C. Limitation and Future Work

Although our proposed method achieves better accuracy than the comparison methods, the model also exhibits some limitations and can be further improved from the following aspects. Deep learning approaches rely heavily on a large amount of high-quality annotations. However, many existing datasets are developed from crowdsource data such as OSM. The ground truth labels are usually not strictly aligned with the image and contain label noise[69], which affects model learning. Exploring noisy label learning to achieve more robust building-road collaborative extraction is one of the future directions. Moreover, generating fine-grained and up-to-date building and road databases from historical[9] or coarse-resolution datasets[70] is worth exploring. Meanwhile, applying transfer learning approaches such as unsupervised domain adaption can improve the model's performance in large-scale mapping[71], [72].

## VI. CONCLUSION

Buildings are the basic carrier of social production and human life; roads are the links that interconnect social networks. The geodata of buildings and roads are valuable for various frontier applications. However, many existing extraction methods ignore the strong spatial correlation between roads and buildings and extract buildings and roads from remote sensing images in isolation. Meanwhile, some multitask methods cause the "seesaw phenomenon" where the accuracy of one task is improved by hurting the performance of another task, which is demonstrated in our experiments. To tackle these problems, we propose a building-road collaborative extraction method by introducing inner-task, cross-task, and cross-scale feature interaction. The features are projected into task-specific and task-shared feature spaces and thus promote cross-task interaction while preserving the information from each task. Furthermore, a cross-scale interaction module is proposed to automatically learn and select the optimal reception field for each feature space, and thus exploit optimal feature representations for each specific task. As compared with the existing methods, our proposed collaborative extraction method can output both the building and road prediction results, improve the inference speed, and enhance the multi-task recognition accuracy through the proposed cross-task and cross-scale feature interaction. Experimental results on two large-scale datasets show that our proposed algorithm can achieve robust and rapid building-road collaborative extraction with strong generalization performance and high extraction accuracy, showing great potential in large-scale mapping.

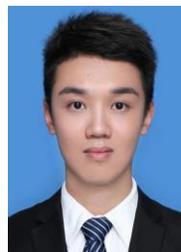

**Haonan Guo** received the B.S. degree in Sun Yat-sen University, Guangzhou, China, in 2020. He is currently pursuing the M.S. degree at the State Key Laboratory of Information Engineering in Surveying, Mapping, and Remote Sensing, Wuhan University. His research interests include deep learning, building footprint extraction, urban remote sensing, and multisensor image processing.

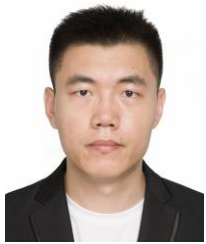

**Xin Su** received the B.S. degree in electronic engineering from Wuhan University, Wuhan, China, in 2008, and the Ph.D. degree in image and signal processing from Telecom ParisTech, Paris, France, in 2015. He was a Post-Doctoral Researcher with the Team SIROCCO, Institut National de Recherche en Informatique et en Automatique (INRIA), Rennes, France. He is currently an Associate Professor with School of Remote Sensing and Information Engineering, Wuhan University, China. His research interests include multi-temporal remote sensing image processing, multiview image processing and 3-D video communication.

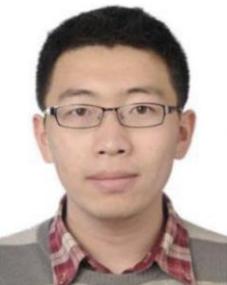

**Chen Wu** (Member, IEEE) received the B.S. degree in surveying and mapping engineering from Southeast University, Nanjing, China, in 2010, and the Ph.D. degree in photogrammetry and remote sensing from the State Key Laboratory of Information Engineering in Surveying, Mapping and Remote Sensing, Wuhan University, Wuhan, China, in 2015.

He is currently a full Professor with the State Key Laboratory of Information Engineering in Surveying, Mapping and Remote Sensing, Wuhan University. His research interests include multitemporal remote sensing image change detection and analysis in multispectral and hyperspectral images.

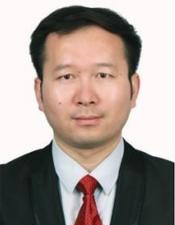

**Bo Du** (Senior Member, IEEE) received the Ph.D. degree in photogrammetry and remote sensing from the State Key Laboratory of Information Engineering in Surveying, Mapping and Remote Sensing, Wuhan University, Wuhan, China, in 2010.

He is a Professor with the School of Computer Science, Wuhan University, and the Institute of Artificial Intelligence, Wuhan University. He is also the Director of the National Engineering Research Center for Multimedia and Software, Wuhan Uni-versity. He has published more than 80 research articles in IEEE TRANSACTIONS ON IMAGE PROCESSING (TIP), IEEE TRANSACTIONS ON CYBERNETICS (TCYB), IEEE TRANSACTIONS ON PAT-TERN ANALYSIS AND MACHINE INTELLIGENCE (TPAMI), IEEE TRANSAC-TIONS ON GEOSCIENCE AND REMOTE SENSING (TGRS), IEEE JOURNAL OF SELECTED TOPICS IN APPLIED EARTH OBSERVATIONS AND REMOTE SENSING (JSTARS), and IEEE GEOSCIENCE AND REMOTE SENSING LET-TERS (GRSL). Thirteen of them are Essential Science Indicators (ESI) hot articles or highly cited articles. His research interests include pattern recognition, hyperspectral image processing, and signal processing.

Dr. Du regularly serves as a Senior Program Committee (PC) Member of the International Joint Conferences on Artificial Intelligence (IJCAI) and the Association for the Advancement of Artificial Intelligence (AAAI). He also serves as a reviewer for 20 Science Citation Index (SCI) magazines, including IEEE TPAMI, TCYB, TGRS, TIP, JSTARS, and GRSL. He received the Highly Cited Researcher by the Web of Science Group in 2019 and 2020, received the IEEE Geoscience and Remote Sensing Society (GRSS) 2020 Transactions Prize Paper Award, received the IJCAI Distinguished Paper Prize, was the IEEE Data Fusion Contest Champion, and received the IEEE Workshop on Hyperspectral Image and Signal Processing Best Paper Award in 2018. He has served as an Area Chair for the International Conference on Pattern Recognition (ICPR). He serves as an Associate Editor for Neural Networks, Pattern Recognition, and Neurocomputing.

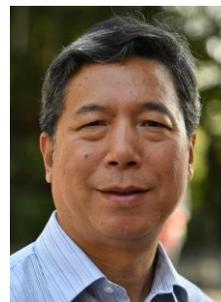

**Liangpei Zhang** (Fellow, IEEE) received the B.S. degree in physics from Hunan Normal University, Changsha, China, in 1982, the M.S. degree in optics from Xi'an Institute of Optics and Precision Mechanics, Chinese Academy of Sciences, Xi'an, China, in 1988, and the Ph.D. degree in photogrammetry and remote sensing from Wuhan University, Wuhan, China, in 1998.

He was a Principal Scientist for the China State Key Basic Research Project (2011–2016) appointed by the Ministry of National Science and Technology of China to lead the remote sensing program in China. He is a Chair Professor with the State Key Laboratory of Information Engineering in Surveying, Mapping, and Remote Sensing (LIESMARS), Wuhan University. He has published more than 700 research articles and five books. He is the Institute for Scientific Information (ISI) Highly Cited Author. He is the holder of 30 patents. His research interests include hyperspectral remote sensing, high-resolution (HR) remote sensing, image processing, and artificial intelligence.

Dr. Zhang is a fellow of the Institution of Engineering and Technology (IET), London, U.K. He was a recipient of the 2010 Best Paper Boeing Award, the 2013 Best Paper ERDAS Award from the American Society of Photogrammetry and Remote Sensing (ASPRS), and the 2016 Best Paper Theoretical Innovation Award from the International Society for Optics and Photonics (SPIE). His research teams won the top three prizes of the IEEE Geoscience and Remote Sensing Society (GRSS) 2014 Data Fusion Contest, and his students have been selected as the winners or finalists of the IEEE International Geoscience and Remote Sensing Symposium (IGARSS) student paper contest in recent years. He is the Founding Chair of IEEE GRSS Wuhan Chapter. He also serves as an associate editor or editor for more than ten international journals. He is currently serving as an Associate Editor for the IEEE TRANSACTIONS ON GEOSCIENCE AND REMOTE SENSING.